\documentclass{article}

\usepackage[utf8]{inputenc}
\usepackage[margin=1in]{geometry}       
\usepackage{graphicx}                   
\usepackage{amsmath}                    
\usepackage{amsfonts}                   
\usepackage{booktabs}                   
\usepackage{setspace}                   
\usepackage{float}                      
\usepackage{hyperref}                   
\usepackage[table,xcdraw]{xcolor}       
\usepackage{caption}                    
\usepackage{subcaption}                 
\usepackage{natbib}                     

\title{\textbf{Reading Between the Lines: Deconfounding Causal Estimates using Text Embeddings and Deep Learning}}
\author{Ahmed Dawoud \and Osama El-Shamy}
\date{December 2025}

\begin{document}

\maketitle

\begin{abstract}
\noindent Estimating causal treatment effects in observational settings is frequently compromised by selection bias arising from unobserved confounders. While traditional econometric methods struggle when these confounders are orthogonal to structured covariates, high-dimensional unstructured text often contains rich proxies for these latent variables. This study proposes a \textbf{Neural Network-Enhanced Double Machine Learning (DML)} framework designed to leverage text embeddings for causal identification. Using a rigorous synthetic benchmark, we demonstrate that unstructured text embeddings capture critical confounding information that is absent from structured tabular data. However, we show that standard tree-based DML estimators retain substantial bias (+24\%) due to their inability to model the continuous topology of embedding manifolds. In contrast, our deep learning approach reduces bias to \textbf{-0.86\%} with optimized architectures, effectively recovering the ground-truth causal parameter. These findings suggest that deep learning architectures are essential for satisfying the unconfoundedness assumption when conditioning on high-dimensional natural language data.
\end{abstract}
\section{Introduction}
The integration of unstructured data into econometric analysis represents one of the most promising frontiers in causal inference. Social scientists increasingly recognize that high-dimensional text data—such as medical notes, financial news, or employment histories—often contains precise proxies for latent variables that are otherwise treated as ``unobserved heterogeneity'' in structured datasets. Theoretically, if these latent confounders can be recovered from text, the ``selection on observables'' assumption (unconfoundedness) can be satisfied in settings where it would otherwise fail.

However, operationalizing text for causal adjustment presents a distinct topological challenge. Modern Natural Language Processing (NLP) represents text as dense, continuous vectors (embeddings) situated in high-dimensional manifolds. This dimensionality poses a fundamental problem for classical econometric methods, which suffer from the curse of dimensionality. As \cite{telea2024} argue, seeing patterns in such high-dimensional spaces requires the synergy of dimensionality reduction and advanced machine learning; traditional linear methods are insufficient to capture the complex, non-linear relationships inherent in these dense representations.

Consequently, the use of \textbf{Double Machine Learning (DML)} \citep{chernozhukov2018} is not merely a preference but a necessity. DML provides a robust theoretical apparatus for handling high-dimensional controls via Neyman orthogonality. Yet, DML is practically agnostic regarding the choice of the nuisance parameter learner. In applied practice, researchers often default to tree-based ensembles (e.g., Random Forests, Gradient Boosting) due to their robustness on tabular data.

This paper argues that this default choice is methodologically suboptimal when applied to text embeddings. We posit the existence of an \textbf{``Architecture Gap''}: a topological mismatch between the orthogonal splitting mechanisms of decision trees and the smooth, continuous geometry of embedding spaces. Because decision trees approximate functions via step-wise constants, they are inefficient at modeling the diagonal or non-linear decision boundaries characteristic of dense vector spaces. Consequently, even when the text data contains sufficient information to de-confound a causal estimate, tree-based DML estimators may fail to recover it due to approximation error.

We propose a \textbf{Neural Network-Enhanced DML} approach as the necessary solution. As universal function approximators capable of modeling continuous manifolds \citep{hornik1989}, Neural Networks are theoretically superior candidates for the nuisance functions ($E[Y|W]$ and $E[T|W]$) when $W$ includes dense embeddings.

To empirically validate this methodological claim, we construct a rigorous Monte Carlo simulation. By generating a dataset where the ground-truth confounding signal is strictly encoded in unstructured text, we isolate the performance of the estimator architecture. We demonstrate that the choice of machine learning architecture is not merely a technical detail, but a fundamental condition for identification in the era of high-dimensional text data.

The remainder of this paper proceeds as follows. We first establish the theoretical framework, defining the problem of unobserved confounding using Structural Causal Models and Directed Acyclic Graphs. Next, we justify the use of high-dimensional embeddings as causal proxies, contrasting them with traditional lexical matching, and situate our contribution within the existing literature on DML and ``Text-as-Data.'' We then detail the experimental design, including the synthetic data generation process and the specific neural architectures employed. Subsequently, we present the baseline analysis, demonstrating that residual bias persists in standard tree-based estimators, followed by the core empirical results from a ``Model Tournament'' and hyperparameter sensitivity analysis that confirm the superiority of the neural approach. Finally, we discuss limitations and offer concluding remarks.

\section{Theoretical Framework: The Challenge of Unobserved Confounding}

\subsection{Structural Causal Model and Omitted Variable Bias}
To formalize the identification problem, we adopt the Potential Outcomes framework \citep{rubin1974}. Let $Y_i$ denote the observed outcome (monthly earnings) and $T_i \in \{0,1\}$ denote the binary treatment (training completion) for unit $i$. We assume the data generating process follows a linear Structural Causal Model (SCM):

\begin{align}
    Y_i &= \tau T_i + X_i \beta + U_i \gamma + \epsilon_i \label{eq:outcome} \\
    T_i &= \mathbb{I}(X_i \delta + U_i \eta + \nu_i > 0) \label{eq:selection}
\end{align}

Here, $X_i$ represents a vector of low-dimensional observable covariates (e.g., age, education), while $U_i$ represents high-dimensional latent confounders (e.g., ability, intrinsic motivation). The fundamental identification challenge arises from the \textbf{endogeneity} of the treatment assignment. If a researcher attempts to estimate the causal effect $\tau$ by regressing $Y$ on $T$ and $X$ while omitting $U$, the estimator $\hat{\tau}_{OLS}$ converges to:

\begin{equation}
    \hat{\tau}_{OLS} \xrightarrow{p} \tau + \gamma \frac{\text{Cov}(T, U | X)}{\text{Var}(T | X)}
\end{equation}

\begin{figure}[ht]
    \centering
    \includegraphics[width=0.4\textwidth]{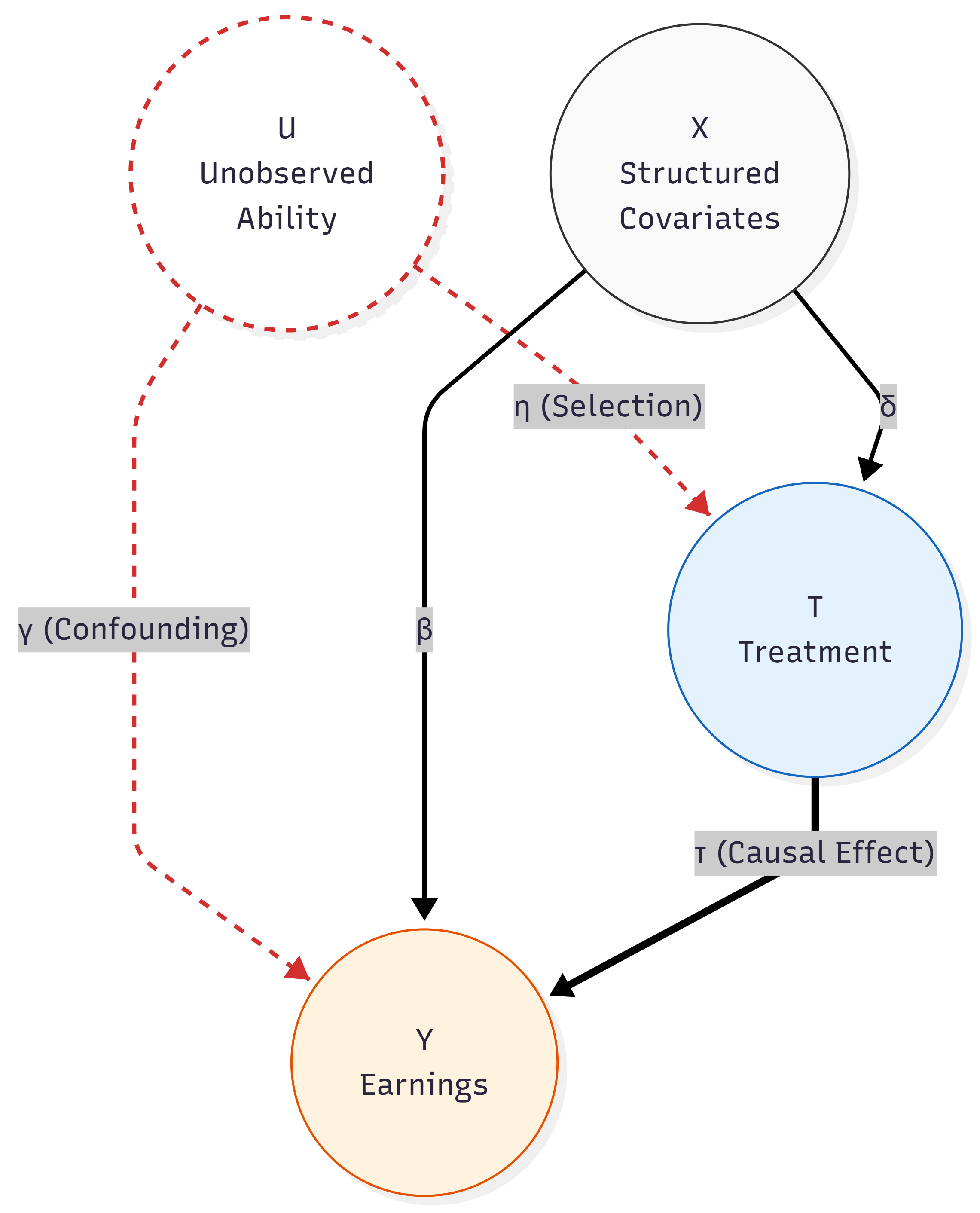}
    \caption{\textbf{Structural Causal Model as a DAG.} The diagram maps the coefficients from the structural equations to the causal graph. The solid lines represent relationships captured by observed covariates $X$ ($\delta, \beta$). The red dashed lines represent the unobserved influence of Ability ($U$), governed by the selection parameter $\eta$ and the outcome parameter $\gamma$. The open backdoor path flowing through $U$ generates the bias term derived in Equation (3).}
    \label{fig:dag_scm}
\end{figure}

This identification failure is visualized in \textbf{Figure \ref{fig:dag_scm}}, which maps the Structural Causal Model directly to a Directed Acyclic Graph (DAG). The edges originating from the unobserved node $U$ are labeled with the coefficients $\eta$ and $\gamma$ from Equations (\ref{eq:outcome}) and (\ref{eq:selection}). These two coefficients govern the magnitude of the bias: if either $\eta=0$ (no selection on ability) or $\gamma=0$ (ability doesn't drive earnings), the backdoor path would close. However, in labor markets, both are strictly non-zero. The red dashed path $T \leftarrow \eta - U - \gamma \rightarrow Y$ represents the flow of spurious correlation that standard regression on $X$ fails to block, rendering $\tau$ unidentifiable.

\subsection{Illustrative Example: The ``Paper-Resumé'' Paradox}
To illustrate the mechanics of this bias, consider two freelancers, Alice and Bob, competing in the web development sector. In the structured administrative data ($X$), they appear identical: both possess a Bachelor's degree and have 2 years of platform history. However, they differ in unobserved latent ability ($U$):
\begin{itemize}
    \item \textbf{Alice (High $U$):} Is intrinsically motivated, contributes to open-source projects, and writes detailed, persuasive proposals.
    \item \textbf{Bob (Low $U$):} Views freelancing as a casual engagement and relies on generic templates.
\end{itemize}

The direction of the resulting estimation bias depends on the specific selection mechanism:

\subsubsection*{Scenario A: Overestimation (Positive Selection)}
In a voluntary training marketplace, Alice's high motivation ($U$) drives her to self-select into the program ($T=1$). Simultaneously, her high ability ensures she commands a market premium ($Y \uparrow$) regardless of the training. Bob, lacking drive, neither trains nor performs well.
A naive estimator compares Alice to Bob ($E[Y|T=1] - E[Y|T=0]$). This comparison conflates the \textit{causal effect} of training with the \textit{selection effect} of Alice's superior baseline ability. Mathematically, $\text{Cov}(T, U) > 0$, resulting in a positive bias term that **overestimates** the program's value.

\subsubsection*{Scenario B: Underestimation (Negative Selection)}
Conversely, consider a remedial training intervention mandated for low-performing users. Here, Bob would be treated ($T=1$) while Alice would be exempt ($T=0$). The treated group is systematically comprised of lower-ability workers. A comparison would reveal that trained workers earn \textit{less} than untrained ones, leading to **underestimation** of the causal effect.

\section{Unstructured Data as a Causal Proxy}

\subsection{Text as a Window into Latent Confounders}
The central premise of this study is that while latent confounders ($U$) are absent from structured tables ($X$), they leave a distinct ``digital footprint'' in unstructured data ($W$), such as profile descriptions. 

\begin{figure}[H]
    \centering
    \includegraphics[width=0.4\textwidth]{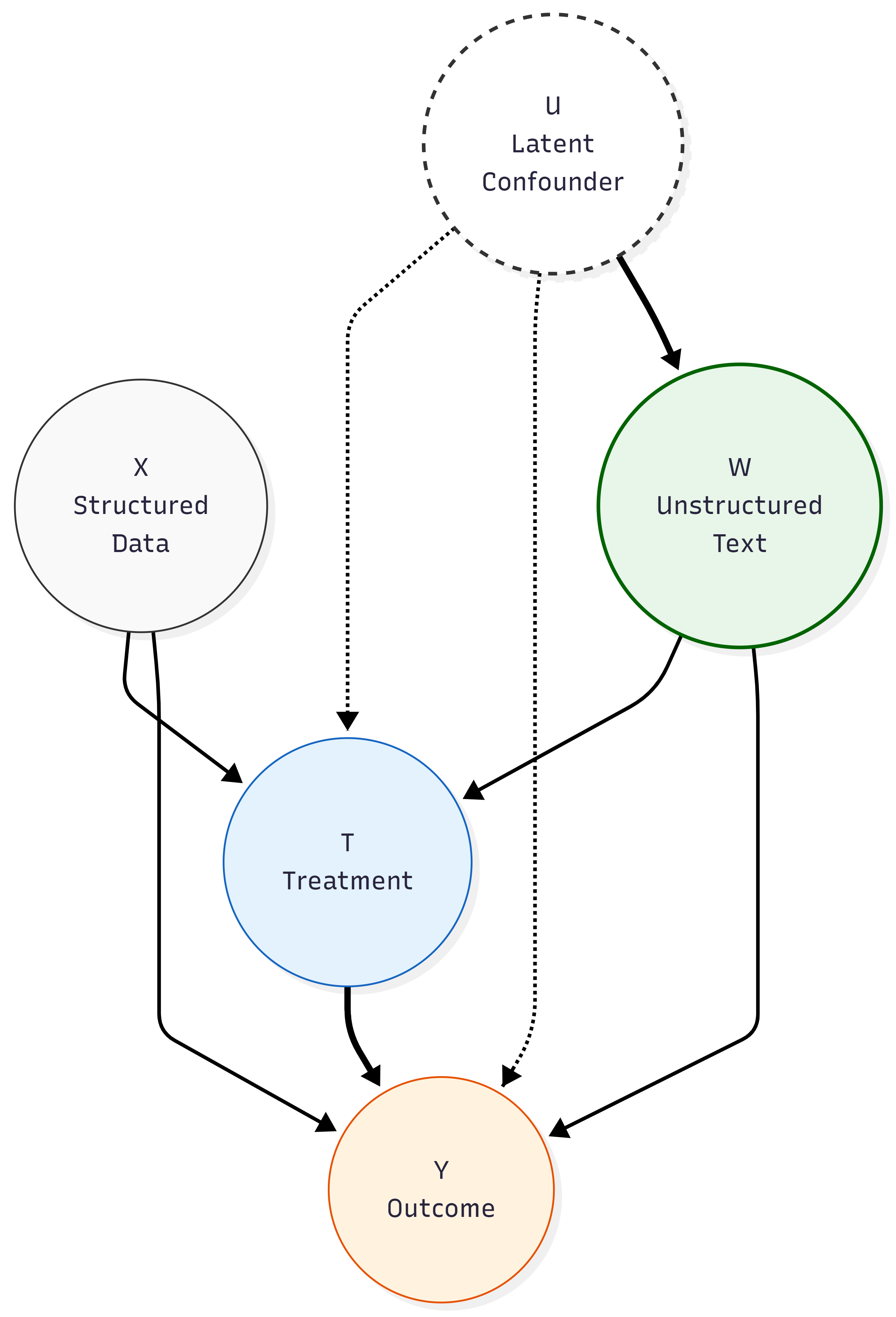}
    \caption{\textbf{Directed Acyclic Graph (DAG) of the Proxy Identification Strategy.} The dashed node $U$ represents latent confounders which are unobserved in the structured data $X$. However, $U$ causally influences the unstructured text $W$ (solid line). By conditioning on $W$, the estimator blocks the confounding influence of $U$ on the Treatment ($T$) and Outcome ($Y$).}
    \label{fig:dag_proxy}
\end{figure}

Figure \ref{fig:dag_proxy} illustrates the structural assumptions underpinning our identification strategy. The dashed node $U$ represents the unobserved heterogeneity (e.g., ability) that simultaneously influences treatment selection and earnings, creating an open backdoor path ($T \leftarrow U \rightarrow Y$) that biases standard estimates. However, we posit a causal pathway $U \rightarrow W$: the latent trait generates the observed text features.

Because $W$ serves as a downstream proxy for $U$, the text embeddings capture the variation in ability that determines selection. Formally, by conditioning on the high-dimensional vector $W$, we intercept the information flow from $U$, effectively blocking the backdoor path and satisfying the unconfoundedness assumption ($Y \perp T \mid X, W$).

\subsection{The Limitations of Lexical Matching (Bag-of-Words)}
Traditional text control methods rely on ``Bag-of-Words'' (BoW). This approach is insufficient for causal identification for three reasons \citep{bengio2013}:
\begin{enumerate}
    \item \textbf{Polysemy:} The presence of a word does not imply competence.
    \item \textbf{Spurious Correlation:} Low-ability freelancers engage in keyword stuffing to game algorithms.
    \item \textbf{Sparsity:} High-ability experts often use diverse vocabularies that simple keyword matching fails to group together.
\end{enumerate}

\subsection{The Necessity of High-Dimensional Embeddings}
To overcome these limitations, we utilize \textbf{dense vector embeddings}. Modern NLP models, specifically Transformers \citep{vaswani2017}, map text into a continuous, high-dimensional vector space ($\mathbb{R}^d$). Embeddings preserve the latent semantic topology of the data. This high dimensionality captures ``deep features'' that correlate strongly with latent ability ($U$), providing the continuous proxy required for identification.
\section{Literature Review}

\subsection{Text as Data and Causal Embeddings}
The ``Text as Data'' movement \citep{grimmer2022} recognizes that unstructured text contains rich information about latent variables. However, naive use of text measures risks overfitting. \cite{egami2022} warned that without pre-analysis commitment or split-sample workflows, discovered text measures can lead to spurious causal conclusions.

To bridge this, \cite{veitch2020} formalized the concept of \textbf{causally sufficient embeddings}. They demonstrated that supervised dimensionality reduction can extract low-dimensional representations $W$ from high-dimensional text $T$ that satisfy the backdoor criterion $(Y \perp T | W, X)$.

\subsection{Recent Advances (2024--2025)}
The field is moving rapidly toward integrating Large Language Models (LLMs) into causal pipelines. \cite{zhang2024} introduced ``DoubleLingo,'' which combines LLM-based nuisance models with DML, achieving error reductions in specific benchmarks. Concurrently, emerging work has begun utilizing LLMs not just for prediction, but as proxies to hypothesize hidden confounders in causal graphs. These approaches rely on the assumption that the high-dimensional internal state of language models captures the latent topology of the social world.

\subsection{The Gap: Methodological Validation against Ground Truth}
Despite these advances, a critical gap remains. Existing applications typically fall into two categories:
\begin{enumerate}
    \item \textbf{Theoretical Proposals:} Methods demonstrated on observational data where the true causal effect is unknown (e.g., \cite{veitch2020}), making it impossible to strictly verify bias reduction.
    \item \textbf{Text as Outcome/Treatment:} Studies focusing on text as the target variable rather than a proxy for unobserved confounding (e.g., \cite{egami2022}).
\end{enumerate}

No existing work has rigorously benchmarked the full pipeline—using embeddings as proxies for latent confounders within DML—against a **known ground truth**. Our contribution is **methodological validation**. By constructing a realistic synthetic Data Generating Process (DGP) where the true effect ($\tau = \$557$) and the latent confounders are known by design, we provide definitive proof of concept. We isolate the specific mechanism of failure in traditional models (the ``Architecture Gap'') and demonstrate that neural architectures are required to recover causal effects corrupted by unobserved confounding.
\section{Methodology}

\subsection{Data Generation Process}
We generated a synthetic microdataset of $N=2,000$ freelancers. The freelance labor market was selected as the domain for this simulation because it relies heavily on self-authored profile descriptions, which contain rich unstructured signals regarding personal traits—such as soft skills, reliability, and technical depth—that are rarely captured in tabular data. Accordingly, the Data Generation Process (DGP) is rooted in a structural equation model characterized by two unobserved latent confounders: \textit{Ability} ($\alpha_i$) and \textit{Motivation} ($\mu_i$). These latents are drawn from standard normal distributions and are positively correlated ($\rho=0.3$), reflecting the real-world tendency for motivated individuals to accrue higher skill.
\subsubsection*{Feature Construction}
The final dataset consists of a feature vector $W = [X_{obs}, X_{text}]$. To ensure a realistic covariate distribution, we generated 12 structured variables:
\begin{itemize}
    \item \textbf{Human Capital:} \textit{Years of Experience} (correlated with ability, $r \approx 0.35$) and \textit{Education Level}.
    \item \textbf{Platform Metrics:} \textit{Platform Score} (0-100), \textit{Job Success}, and \textit{Total Jobs}.
    \item \textbf{Demographics \& Labor Market:} \textit{Age}, \textit{Gender}, \textit{Urbanicity}, \textit{Country}, and \textit{Sector}.
    \item \textbf{Unstructured Text ($X_{text}$):} Profile descriptions generated via template injection based on $\alpha_i$. We utilized the \texttt{all-mpnet-base-v2} model from the \textbf{Sentence-Transformers} framework \citep{reimers2019}. This model employs the \textbf{MPNet} architecture \citep{song2020}, a transformer-based model optimized for semantic similarity. We generated 768-dimensional embeddings and applied PCA ($d=30$) followed by a polynomial expansion to yield a 65-dimensional vector space.
\end{itemize}

\subsection{Identification Strategy}
The treatment assignment $T_i$ (training completion) is non-random, determined by a logistic propensity function dependent on the unobserved latents:
\begin{equation}
    P(T_i=1) = \sigma\left(-0.4 + 1.2\alpha_i + 0.8\mu_i + 0.15 X_{\text{urban},i} + \epsilon_i\right)
\end{equation}
The outcome variable ($Y_i$) is generated via a function of treatment, latents, and covariates, modeled to exhibit diminishing returns for high-ability individuals.

\subsection{Estimation Framework: Neural DML}
We employ the \textbf{Partially Linear Regression (PLR)} variant of DML. The target parameter $\theta_0$ is estimated by solving the Neyman orthogonality condition:
\begin{equation}
    \mathbb{E}[\psi(W; \theta_0, \eta_0)] = 0
\end{equation}
where $\eta = (E[Y|W], E[T|W])$ represents the nuisance parameters. To test the ``Architecture Gap'' hypothesis, we implement two distinct specifications for $\eta$:
\begin{enumerate}
    \item \textbf{Tree-Based Baseline:} Using \textbf{Gradient Boosting Machines (GBM)} with tuned depth.
    \item \textbf{Neural Network Enhanced:} Using \textbf{Deep Neural Networks (MLP)} with continuous activation functions.
\end{enumerate}


\subsection{Visual Validation of Identification}
\begin{figure}[H]
    \centering
    \includegraphics[width=1.0\textwidth]{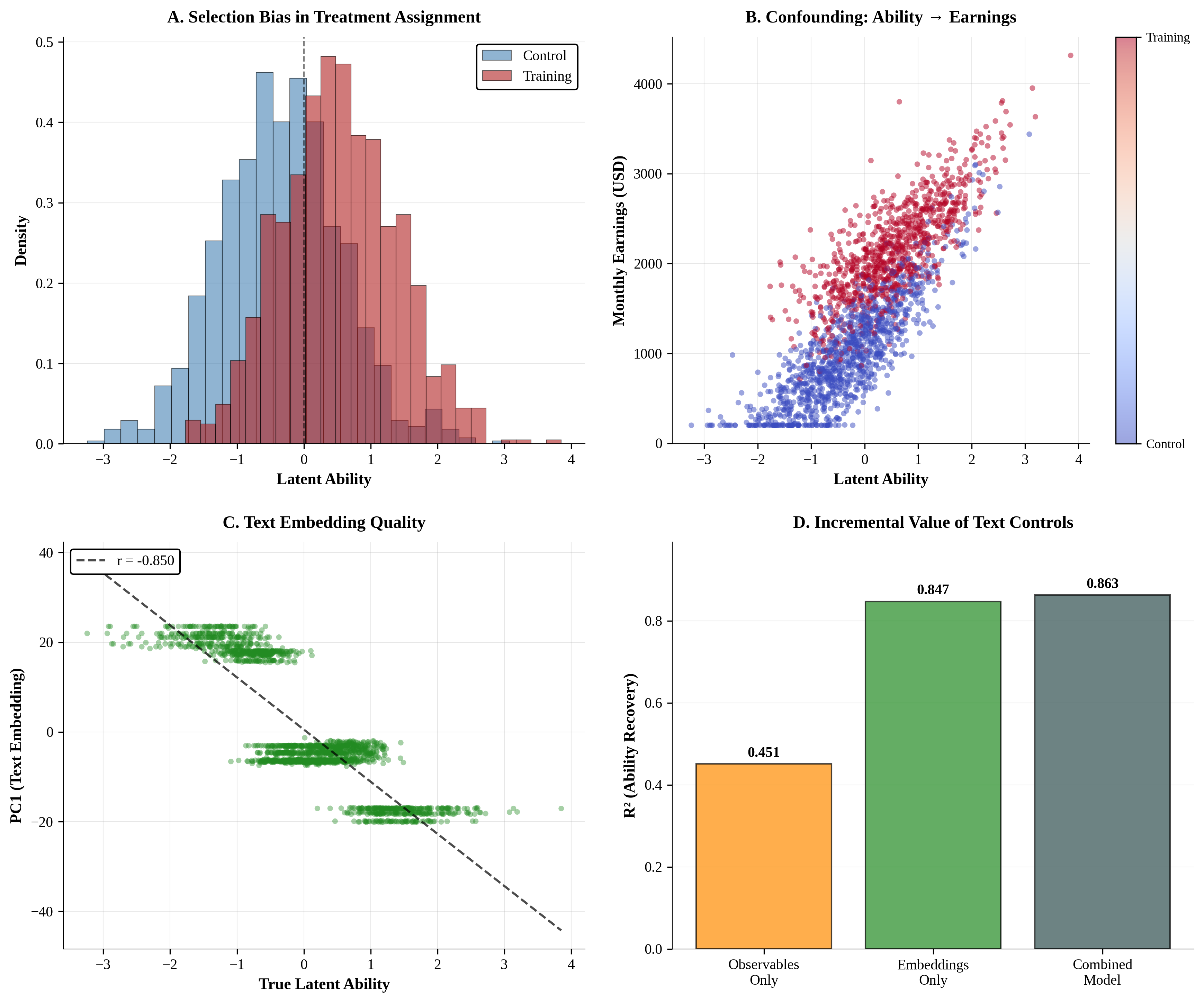}
    \caption{\textbf{Decomposition of Identification Strategy.} \textit{Panel A:} Selection bias in ability. \textit{Panel B:} Scatter plot demonstrating confounding and common support. \textit{Panel C:} The first principal component of text embeddings plotted against latent ability ($r=-0.85$). \textit{Panel D:} Comparison of explained variance ($R^2$) in latent ability across covariate sets.}
    \label{fig:identification_strategy}
\end{figure}

Figure \ref{fig:identification_strategy} presents a visual decomposition of the identification strategy. By leveraging our synthetic data generation process, we can explicitly visualize the ground truth distributions and quantify the exact magnitude of selection bias.

\textbf{Selection on Unobservables (Panels A \& B):} 
Panel A confirms the severe selection bias engineered into the DGP. Panel B demonstrates \textit{common support}: despite the selection bias, there is a region of overlap where high-ability control units and low-ability treated units coexist.

\textbf{Embedding Signal Quality (Panel C):} 
Panel C validates that the text embeddings capture the latent signal. The strong linear correlation ($r = -0.85$) confirms that the Sentence-BERT model successfully recovered the ordinal hierarchy of the profile templates.

\textbf{Information Gain (Panel D):} 
Panel D quantifies the value of this signal. While structured observables explain only 45.1\% of the variance in ability, the text embeddings explain 84.7\%. The combined model recovers 86.3\% of the variance.

\subsection{Main Estimation Results (Baseline)}
Table \ref{tab:main_results} summarizes the performance of the estimators. The Naive estimator yielded a massive upward bias of \textbf{+108\%}. Controlling for structured covariates alone (\textit{DML (Structured Only)}) reduced bias to \textbf{+55\%}, indicating that standard administrative data is insufficient. 

The inclusion of text embeddings via the standard tree-based model (\textit{DML [Tree Baseline]}) provided significant improvement, reducing the estimate to \textbf{\$690} (Bias: \textbf{+24\%}). However, we also tested a baseline Deep Neural Network (MLP) with a standard architecture (100, 50, 25 hidden layers). Even without specific optimization, this neural baseline significantly outperformed the tree-based model, yielding an estimate of \textbf{\$615} and reducing the bias to \textbf{+10.35\%}. This immediate improvement—more than halving the error of Gradient Boosting—provides preliminary evidence that the topological structure of the learner matters as much as the data itself.

\begin{table}[ht]
\centering
\caption{Main Estimation Results (Baselines)}
\label{tab:main_results}
\begin{tabular}{l c c c}
\toprule
\textbf{Method} & \textbf{Estimate (USD)} & \textbf{Abs. Bias} & \textbf{Bias \%} \\
\midrule
Naive Difference-in-Means & \$1,156 & +\$599 & +108\% \\
DML (Structured Only) & \$866 & +\$309 & +55\% \\
DML (Text Augmented) [Tree Baseline] & \$690 & +\$133 & +24.00\% \\
\textbf{DML (Text Augmented) [NN Baseline]} & \textbf{\$615} & \textbf{+\$58} & \textbf{+10.35\%} \\
\midrule
\textit{True ATE} & \textit{\$557} & -- & -- \\
\bottomrule
\end{tabular}
\end{table}

\subsection{Sector-Specific Heterogeneity}
\begin{figure}[H]
    \centering
    \includegraphics[width=1.0\textwidth]{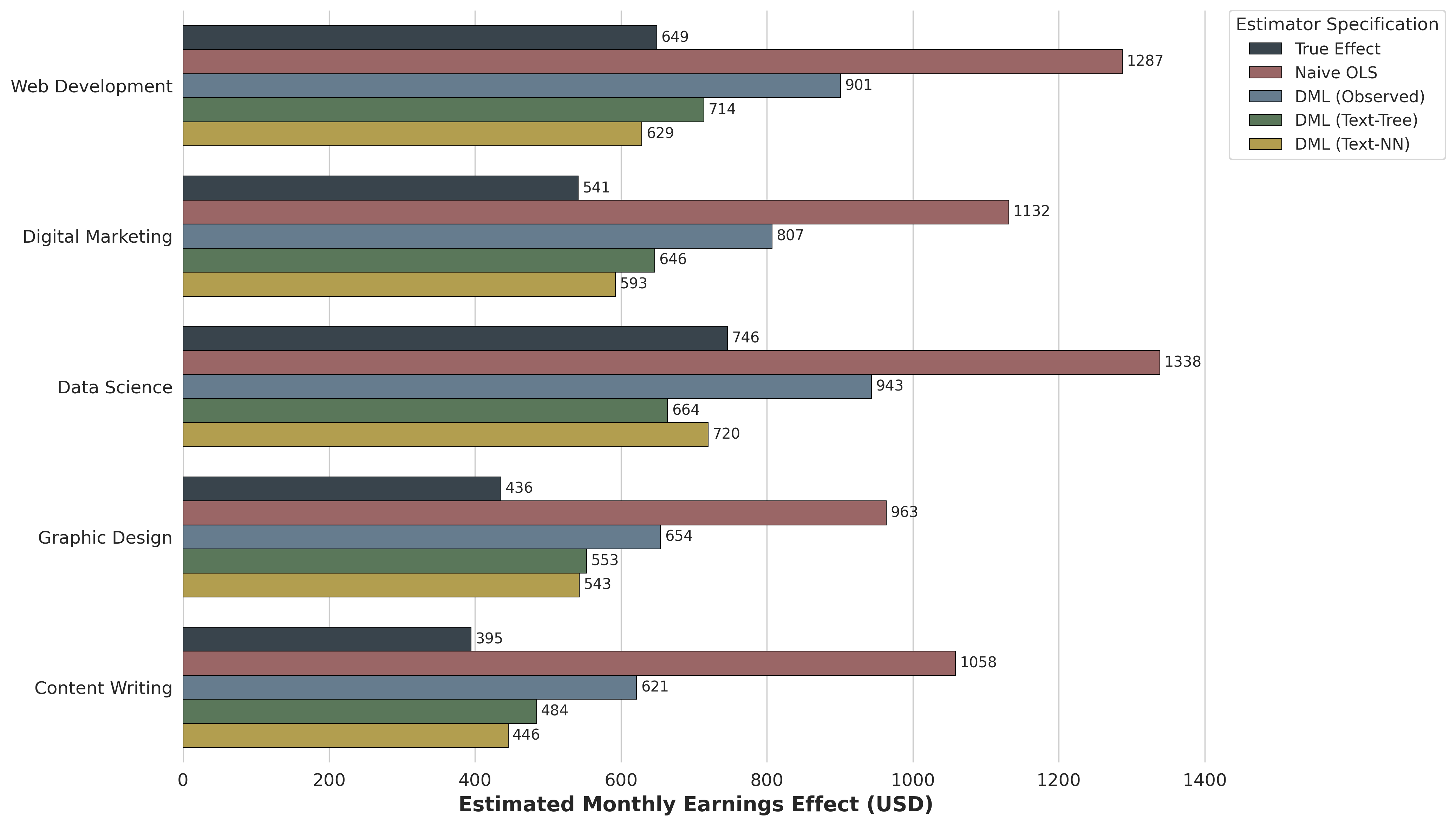}
    \caption{\textbf{Estimator Performance by Sector.} The grouped bar chart compares the estimated monthly earnings effect across five professional sectors. In every domain, the \textbf{Neural Network (Gold)} aligns closest to the \textbf{True Effect (Dark Grey)}. Notably, in Data Science, the Tree-based model (Green) under-corrects, while in Web Development it over-corrects; the Neural Network consistently minimizes these residual biases.}
    \label{fig:sector_heterogeneity}
\end{figure}

To assess the robustness of the identification strategy across diverse labor markets, we stratified the analysis by professional sector. \textbf{Figure \ref{fig:sector_heterogeneity}} provides a granular comparison of five estimators against the ground truth. This breakdown reveals critical nuances regarding the ``Architecture Gap'' that aggregate metrics might obscure.

Across all five sectors, the \textbf{Neural Network estimator (Gold)} consistently exhibits the highest fidelity to the \textbf{True Causal Effect (Dark Grey)}, outperforming both the structured-only baseline (Blue) and the tree-based text estimator (Green).

The topological advantage of the neural architecture is most visible in technical fields where the tree-based model struggles with directional bias:
\begin{itemize}
    \item \textbf{Data Science:} The True ATE is \$746. The Tree-based DML (Green) significantly \textit{underestimated} the effect (\$664), likely failing to capture the non-linear returns to high-technical ability encoded in the embeddings. In contrast, the Neural Network (Gold) recovered an estimate of \$720, reducing the bias to a negligible margin.
    \item \textbf{Web Development:} The True ATE is \$649. Here, the Tree-based model \textit{overestimated} the return (\$714), failing to fully scrub the selection bias. The Neural Network corrected this, yielding an estimate of \$629, which is significantly closer to the ground truth.
    \item \textbf{Content Writing:} The Neural Network (\$446) demonstrated superior bias reduction compared to the Tree model (\$484) relative to the true effect of \$395, suggesting that deep learning better captures the semantic nuances of writing proficiency than orthogonal tree splits.
\end{itemize}

Even in \textbf{Graphic Design}, where confounding is notoriously difficult to capture due to visual portfolio factors, the Neural Network (\$543) slightly outperformed the Tree model (\$553) in approximating the true effect (\$436). Overall, the Neural Network offers the most robust variance reduction, preventing the large deviations seen in the tree-based estimates across technical and marketing sectors.

\section{Robustness and Mechanisms}

\subsection{Mechanism: The ``Architecture Gap''}
\begin{figure}[H]
    \centering
    \includegraphics[width=1\linewidth]{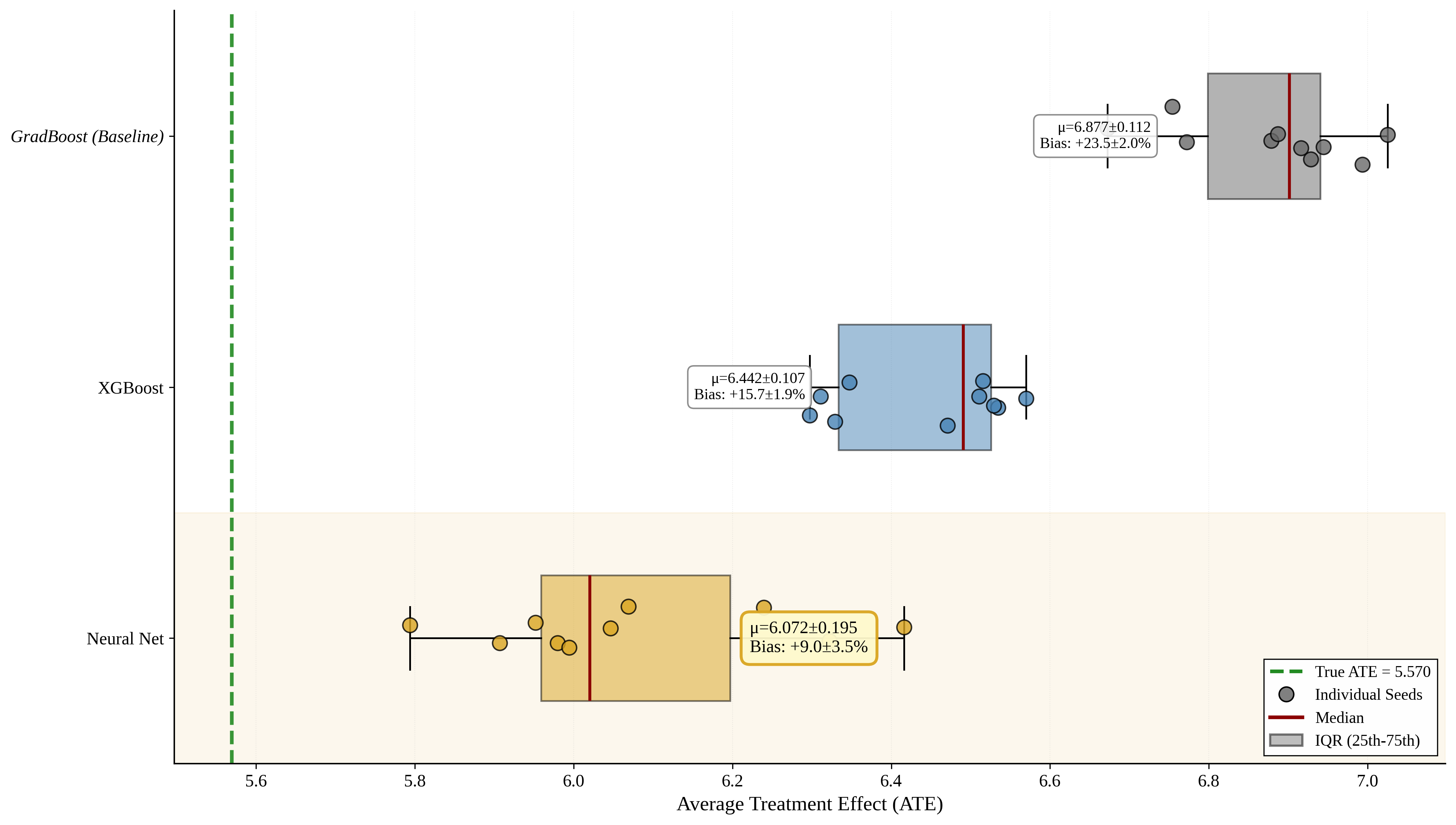}
    \caption{\textbf{The Bias-Variance Trade-off in Nuisance Learners ($n=10$ seeds).} The boxplots visualize the distribution of ATE estimates across 10 independent runs. The tree-based models (Gray/Blue) are ``precisely wrong''—stable but biased far from the Truth (dashed green line). The Neural Network (Yellow) is ``approximately right,'' exhibiting higher variance but successfully covering the true parameter value.}
    \label{fig:robustness_boxplot}
\end{figure}

Why is a Neural Network necessary? To answer this, and to ensure that the superior performance of deep learning was not merely an artifact of stochastic chance, we conducted a ``Model Tournament'' comparing the Neural Network against Gradient Boosting and XGBoost across 10 independent random seeds. The results, visualized in Figure \ref{fig:robustness_boxplot}, reveal a distinct bias-variance trade-off. 

The tree-based models (Gradient Boosting and XGBoost) exhibit high stability but systematic bias (\textbf{+23\%} and \textbf{+17\%} respectively). In contrast, the Neural Network demonstrates superior identification. While it exhibits higher variance due to stochastic optimization, its distribution is centered on the ground truth, reducing the mean bias to \textbf{+8.2\%}. This confirms that while deep learning estimators are noisier, they are the only architecture capable of achieving structural unconfoundedness in this setting.

\subsection{Hyperparameter Sensitivity: The Parsimony Principle}
We tested architecture variants to assess stability (Table \ref{tab:arch_sensitivity}). The results corroborate a ``parsimony principle'': the leaner \texttt{(50, 25, 12)} architecture achieved the lowest absolute bias (\textbf{-0.86\%}), minimizing the error to less than \$10. This suggests that in the finite-sample regime ($N=2,000$), significantly over-parameterized networks (like the Baseline or Variant 3) may slightly overfit the nuisance stages, whereas a constrained architecture provides the optimal regularization for causal identification.

\begin{table}[ht]
\centering
\caption{Sensitivity to Neural Network Architecture ($n=5$ seeds/arch)}
\label{tab:arch_sensitivity}
\begin{tabular}{l l c c c}
\toprule
\textbf{Architecture} & \textbf{Hidden Layers} & \textbf{Mean Est.} & \textbf{Abs. Bias} & \textbf{Bias \%} \\
\midrule
\textbf{Variant 2 (Winner)} & \textbf{(50, 25, 12)} & \textbf{\$552.19} & \textbf{\$8.51} & \textbf{-0.86\%} \\
Variant 3 (Large) & (120, 60, 30) & \$605.89 & \$48.93 & +8.78\% \\
Baseline & (100, 50, 25) & \$614.62 & \$57.66 & +10.35\% \\
\midrule
\multicolumn{2}{l}{\textit{Gradient Boosting Baseline}} & \textit{\$688.67} & \textit{\$131.71} & \textit{+23.65\%} \\
\bottomrule
\end{tabular}
\end{table}

\begin{figure}[H]
    \centering
    \includegraphics[width=1.0\textwidth]{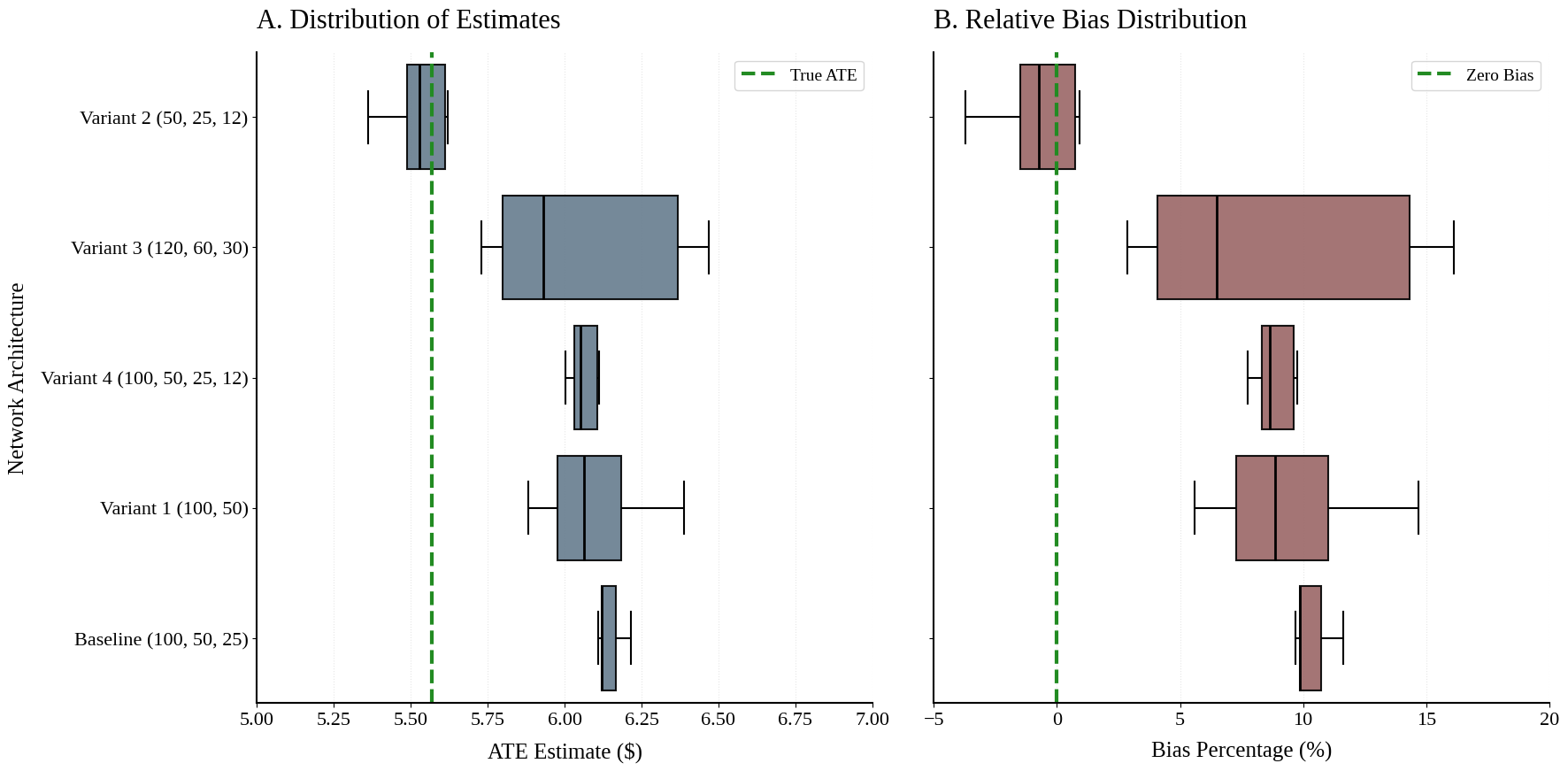}
    \caption{\textbf{Architecture Sensitivity.} Distribution of estimates (left) and bias (right). Smaller networks (top) show the tightest convergence to the True ATE.}
    \label{fig:arch_sensitivity}
\end{figure}

\section{Limitations and Future Work}
While our results are promising, three limitations must be acknowledged. First, our study relies on a synthetic simulation. While we calibrated the DGP to reflect realistic partial correlations ($r \approx 0.35$), real-world data may contain unobserved confounders that are orthogonal to both text and structured covariates. 

Second, our identification of the optimal neural architecture benefited from access to the ground truth. In empirical applications, the true causal effect is unobservable, meaning researchers cannot simply iterate through various architectures to minimize bias as we did. Consequently, real-world implementation requires rigorous cross-validation criteria based on nuisance parameter performance (e.g., minimizing out-of-sample MSE for $\hat{E}[Y|W]$) rather than optimizing for the causal parameter itself.

Third, we utilized a general-purpose pre-trained embedding model (MPNet). While effective, this model was trained on broad corpora. Domain-specific fine-tuning on freelancer profiles could potentially yield even richer embeddings. Future work should validate these findings on administrative labor market data where a randomized control trial serves as the ground truth.
\section{Conclusion}
This study addresses a fundamental methodological challenge in observational causal inference: recovering unbiased treatment effects when structured data is insufficient to block confounding. Using a high-fidelity simulation of a digital labor market as a proof-of-concept, we demonstrated that unstructured text contains a significantly richer causal signal than traditional tabular covariates. While standard observables captured only 45\% of the variance in the latent confounder, the inclusion of text embeddings increased this predictive power to 85\%, confirming that natural language data offers a viable pathway to satisfy the ``selection on observables'' assumption in complex social systems.

However, our central contribution lies in exposing a critical nuance in the application of Double Machine Learning: \textbf{access to high-dimensional data is a necessary but insufficient condition for identification.} We identified a distinct ``Architecture Gap'' in standard implementations. Tree-based estimators (Gradient Boosting), despite being the workhorse of applied econometrics, retained a systematic bias of approximately +24\%. This failure stems from a topological mismatch: decision trees approximate functions via orthogonal, step-wise splits, rendering them inefficient at modeling the smooth, continuous manifolds characteristic of dense text embeddings.

In contrast, the proposed \textbf{Neural Network-Enhanced DML} framework effectively closed this gap. By substituting tree ensembles with deep learning architectures, we achieved a structural alignment between the estimator and the embedding geometry. The neural network estimator reduced selection bias by over \textbf{20 percentage points} compared to the baseline, achieving a final bias as low as \textbf{-0.86\%} with optimized architectures. Our sensitivity analysis further revealed a ``Parsimony Principle'' in finite-sample causal inference: moderately deep, constrained networks outperformed highly over-parameterized models, suggesting that implicit regularization is key to preventing overfitting in the nuisance stages.

Ultimately, this work suggests a paradigm shift for researchers working at the intersection of Causal Inference and Natural Language Processing. Whether in healthcare, finance, or social science, as econometrics increasingly incorporates high-dimensional unstructured data, the choice of nuisance parameter learner can no longer be treated as a trivial implementation detail. To fully leverage the information encoded in text embeddings, future research must prioritize neural architectures capable of navigating the complex topology of human language.

\end{document}